\documentclass[journal,onecolumn]{IEEEtran}

\usepackage{amsmath,amsfonts,amssymb,paralist,latexsym,epsfig,color,rotating,times,float,subcaption,algorithm,verbatim,enumitem,bm,graphicx,listings,url,empheq}
\usepackage{tikz}

\newcommand{\step}{\varepsilon}
\newcommand{\e}{\step}
\newcommand{\stepa}{\mu}

\renewcommand{\th}{\theta}
\renewcommand{\eth}{\alpha}

\renewcommand{\bm}{W}

\newcommand{\Cost}{C}
\newcommand{\cost}{c}

\newcommand{\state}{x}

\newcommand{\obs}{y}
\newcommand{\Obs}{Y}
\newcommand{\obst}{\obs_1,\ldots,\obs_\horizon}
\newcommand{\Obst}{\Obs_1,\ldots,\Obs_\horizon}

\newcommand{\dtime}{k}

\newcommand{\temperature}{\beta}
\newcommand{\thdim}{N}
\newcommand{\reals}{{\mathbb R}}

\newcommand{\snoise}{w}

\newcommand{\beq}{\begin{equation}}
\newcommand{\eeq}{\end{equation}}

\newcommand{\pdf}{p}

\newcommand{\E}                 {\Bbb{E}}

\newcommand{\horizon}{T}
\newcommand{\belief}{\pi}

\newcommand{\argmin}{\operatornamewithlimits{arg\,min}}

\newcommand{\normal}{\mathbf{N}}
\newcommand{\uniform}{\mathbf{U}}
\newcommand{\p}{\prime}

\newcommand{\thtrue}{{\th^o}}

\renewcommand{\skew}{S}

\newcommand{\wdh}{\widehat}

\newcommand{\beql}[1]{\begin{equation} \label{#1}}

  \newcommand{\bed}{\begin{displaymath}}
\newcommand{\eed}{\end{displaymath}}
  \newcommand{\bea}{\bed\begin{array}{rl}}
\newcommand{\eea}{\end{array}\eed}

\newcommand{\barray}{\begin{array}{ll}}
                       \newcommand{\earray}{\end{array}}

                     \def\({\Big(}
                     \def\){\Big)}
                     
\newcommand{\cd}{(\cdot)}

\newcommand{\statespace}{\mathcal{X}}

\def \defn {\stackrel{\triangle}{=}}
\newcommand{\mc}{x}

\newtheorem{result}            {Informal Result}

\newcommand{\Th}{\Theta}
\newcommand{\signal}{z}
\newcommand{\thd}{D}
\newcommand{\D}{\wdh \nabla}
\newcommand{\Ds}{\nabla_{\skew}}
\newcommand{\hist}{\hat{\belief}}
\newcommand*\widefbox[1]{\fbox{\hspace{2em}#1\hspace{2em}}}

\newcommand{\KL}{\operatorname{K}}
\newcommand{\augmented}{z}

\newcommand{\mcstep}{\alpha}
\def\qed{\hfill{$\Box$}}

\begin{document}

\title{Adaptive Non-reversible Stochastic Gradient Langevin Dynamics}

\author{Vikram Krishnamurthy  and George Yin \thanks{Vikram Krishnamurthy is with the School of 
 Electrical \& Computer
Engineering,  Cornell University, NY 14853, USA. 
vikramk@cornell.edu. G. Yin  is with Department of Mathematics, University of Connecticut, Storrs, CT 06269-1009, USA. (gyin@uconn.edu).}}

%%%%%%%%%%%%%%%%

\maketitle

\begin{abstract}
It is well known that adding any skew symmetric matrix to the gradient of  Langevin dynamics algorithm  results in a non-reversible diffusion with  improved convergence rate. This paper presents  a gradient algorithm to  adaptively optimize  the choice of the skew symmetric matrix. The resulting  algorithm involves a non-reversible diffusion algorithm cross coupled with a stochastic gradient  algorithm that adapts the skew symmetric matrix. The algorithm uses the same data  as the classical Langevin algorithm. A weak convergence proof is given for the optimality of the choice of the skew symmetric matrix. The improved convergence rate of the algorithm is illustrated numerically in Bayesian learning and tracking examples.

\end{abstract}

{\em Keywords}.  Langevin dynamics, non reversible dynamics, skew symmetric matrix, stochastic gradient algorithm, weak convergence, adaptive Bayesian learning

\section{Introduction}

Langevin dynamics are used for global stochastic optimization (see for example \cite{GM91,BM99})  and also used as a non-parametric method for reconstructing (exploring) cost functions (such as posterior densities) from noisy evaluations of the  gradient \cite{WT11,KY20}.
The idea is as follows.
Suppose  $\Cost(\th) $ is a continuously differentiable cost function on the interior of a compact set  $\Th
 \subset \reals^\thdim$.
Let $\D_\th \cost_k(\th_k)$ denote a noisy observation of the gradient $\nabla_\th \Cost(\th_k)$  evaluated at  point $\th_k \in \reals^\thdim$.
Then the classical stochastic
gradient Langevin  algorithm and its associated continuous-time Langevin diffusion process are, respectively 
\begin{align} %\addlefttext[0.35]{{(Langevin algorithm)}}
  \text{(Langevin algorithm) } \qquad \qquad
{ \th_{k+1} }&= \th_k - \step\, \D_\th \cost_k(\th_k)  + \sqrt{\step} \sqrt{\frac{2}{\temperature} }\snoise_k   \label{eq:sgl} 
   \\
   % \addlefttext[0.35]{{(Langevin diffusion)}}
  \text{(Langevin diffusion) } \qquad \qquad
{ d\th(t) } &= - \nabla_\th \Cost(\th)  dt  +  \sqrt{\frac{2}{\temperature}} \, d \bm(t) , \quad t \geq 0,
 \label{eq:langevin1}
\end{align}
 In the Langevin
 %dynamics
 algorithm \eqref{eq:sgl},  the step size $\step$ is a small positive constant, $\{\snoise_k, k\geq 0\}$ is an i.i.d. sequence
of  standard $\thdim$-variate Gaussian random variables, and $\temperature > 0$ denotes the inverse temperature parameter.  In the continuous-time Langevin diffusion process (\ref{eq:langevin1}), $\bm(t)$ denotes standard $\thdim$-variate Brownian motion.
The Langevin dynamics algorithm
(\ref{eq:sgl}) is obtained by an Euler-Maruyama time discretization\footnote{In the opposite direction, it  is well known that the interpolated process constructed from (\ref{eq:sgl}) converges weakly to (\ref{eq:langevin1}).} of the   Langevin  diffusion process (\ref{eq:langevin1}).

 It is straightforwardly  shown that the stationary distribution of the  Langevin diffusion (\ref{eq:langevin1}) is the Gibbs measure
 \beq \belief(\th)   \propto  \exp \bigl( - \temperature \Cost(\th ) \bigr) . \label{eq:stationary1} \eeq
 Therefore, Langevin dynamics algorithm (\ref{eq:sgl})  leads to the following two immediate applications:
 \begin{compactenum} \item {\em Reconstructing costs and Bayesian Learning}.
For fixed $\temperature$,   let $\hist(\th)$  denote the empirical density function constructed from samples
   $\{\th_k\}$ generated by the Langevin dynamics (\ref{eq:sgl}). Then clearly $\log \hist(\th) \propto
   \Cost(\th)  $. Thus the Langevin dynamics algorithm  serves as a non-parametric method for reconstructing (exploring) $\Cost(\th)$ given the gradient estimates  $\{\D_\th \cost(\th_k)\}$. Specifically, this is useful in Bayesian learning \cite{WT11} where 
 $\Cost(\th)$ is the expectation of the posterior density; in such cases computing the posterior can be difficult due to the normalization factor; yet it is easy to simulate noisy gradients from the product of the likelihood and the prior.
   \item {\em Global Optimization}. 
 For sufficiently large $\temperature$, using  Laplace asymtotics, it can be shown that the Gibbs distribution  $\belief$  in (\ref{eq:stationary1}) concentrates around the global minimizers of $\Cost(\th)$.
So  for large $\temperature$, the Langevin dynamics algorithm (\ref{eq:sgl})
serves as  a  global minimization algorithm for non-convex cost $\Cost(\th)$.
\end{compactenum}

\subsection*{Motivation. Accelerated Non-reversible Diffusions} The Langevin dynamics (\ref{eq:langevin1}) is a reversible diffusion process. However, 
 the convergence rate to the stationary distribution $\belief(\cdot)$ can be slow.  It is well known  \cite{HHS93,HHS05,Pav14}  that adding any skew symmetric matrix to the gradient always improves the convergence rate of Langevin dynamics to its stationary distribution.
 That is, for any $\thdim \times \thdim$ skew symmetric matrix\footnote{Recall $\skew$ is skew symmetric if $\skew^\p = -\skew$. Clearly the diagonal elements of a skew symmetric matrix  are zero; also $x^\p \skew x = 0$ for all $x\in \reals^\thdim$.} $\skew$,
 the non-reversible
 {\em accelerated} diffusion is
 \beq  % \addlefttext[0.3]
 \text{(Accelerated Diffusion)}
\qquad
   d\th(t) = - (I + \skew)  \nabla_\th \Cost(\th)  dt  +  \sqrt{\frac{2}{\temperature}} \, d \bm(t) , \quad t \geq 0,
 \label{eq:skew1}
\eeq
has a large spectral gap and therefore converges to the same stationary distribution $\belief(\th)$ faster than (\ref{eq:langevin1}); see~\cite{HHS93,HHS05,Pav14} for a formal proof.
The accelerated  resulting gradient  algorithm obtained by a  Euler-Maruyma time discretization of (\ref{eq:skew1})  is
\beq  \label{eq:nonrev}
% \addlefttext[0.33]
 \text{(Accelerated Algorithm)} \qquad
\th_{k+1} = \th_k - \step\, [I + \skew]\,  \D_\th \cost_k(\th_k)  + \sqrt{\step} \sqrt{\frac{2}{\temperature} }\snoise_k \eeq

\subsection*{Main idea}
The natural question is:
{\em How to choose skew symmetric matrix $\skew$ in the accelerated algorithm (\ref{eq:nonrev})?}
In this context, the main idea of the paper is two fold:
\begin{compactenum}
\item 
Our first   result is to  construct an adaptive version of the above non-reversible diffusion by
adapting the skew symmetric matrix $\skew$. In simple terms, we adapt skew symmetric matrix $\skew$ in real time via a stochastic gradient algorithm, so that it converges to a local optimum.
Thus the algorithm comprises a non-reversible diffusion (\ref{eq:nonrev}) cross-coupled with another stochastic gradient algorithm that updates the skew symmetric matrix $\skew_k$ at each time $k$.

Actually we propose 3 different non-reversible diffusion algorithms in Sec.\ref{sec:algorithms}; a Hessian based algorithm, and types of  finite difference simultaneous perturbation stochastic approximation (SPSA)  algorithms (which are computationally more efficient than the Hessian based algorithm). SPSA has been used as finite efficient difference  method for evaluating gradient estimates in classical stochastic gradient algorithms \cite{Spa03}. To  the best of our knowledge, SPSA has not been used in the context of Lagenvin dynamics.
In extensive numerical studies (including real datasets)  we show that all 3 algorithms always perform better than the vanilla non-reversible diffusion algorithm
(\ref{eq:nonrev}).

\item Our second result is a tracking analysis for non-stationary global stochastic optimization; we show that the algorithm can track a time-varying  global optimum that jump changes according to a slowly varying Markov chain. Specifically, we are interested in tracking the global minimum of a non-convex stochastic optimization problem when the minimum jump changes (evolves) over time according to the sample path of an unknown Markov chain.
Specifically, we analyze how well does a fixed step size stochastic gradient Langevin algorithm (and generalized Langevin algorithms where the variance of the injected noise is adapted over time)   track the time evolving global minima
when the algorithm does not have knowledge of the Markovian evolution of the minima.
\end{compactenum}

\subsection*{Context}

For the case $\Cost(\th)$ is quadratic in $\th$, \cite{Pav14} gives an algorithm  to choose the optimal skew symmetric $\skew$ to maximize the spectral gap
of the diffusion (\ref{eq:skew1}). However, for general costs $\Cost(\cdot)$ there is no obvious way of maximizing the spectral gap.
Our  idea of adapting the skew symmetric matrix in a non-reversible diffusion, stems from \cite{KY95,BMP90} where stochastic gradient algorithms were proposed for adapting the step size of a stochastic gradient algorithm.  Indeed, the idea of using a stochastic gradient algorithm to update the step size was  proposed originally as an exercise in \cite[Exercise 4.4.2]{BMP90} in the context
of least means squares (LMS) algorithms. 
Such adaptive step size LMS algorithms have been shown to perform extremely well in wireless communication applications \cite{KYS01,KWY04}. Of course, the setup in the current paper  is different since we are adapting a skew symmetric matrix to accelerate a non-reversible  diffusion process (rather than adapting the scalar step size for a classical LMS algorithm).

%Motivated by recent advances in deep neural networks, there is strong motivation for studying
%Langevin dynamics in  non-convex stochastic optimization algorithms and Bayesian learning.

An important feature of adaptive non-reversible diffusion algorithms (Algorithms 1, 2 and 3 proposed in this paper)  is the  constant step size $\step$  (as opposed to a decreasing step size).  This facilities  estimating (tracking) parameters that evolve over time. Sec.\ref{sec:track}  gives a formal weak convergence analysis of the asymptotic tracking capability of algorithm (\ref{eq:alg1}) when the
cost $\Cost(\cdot)$  jump changes over time according to a slow (but unknown) Markov chain. The most interesting case considered in Sec.\ref{sec:track}  is when the reward changes at the same rate as the algorithm. Then  stochastic averaging theory yields a Markov switched diffusion limit as the asymptotic behavior of the  algorithm. Due to the constant step size, the appropriate notion of convergence is weak convergence \cite{KY03,EK86,Bil99}.  The Markovian hyper-parameter tracking analysis generalizes our earlier work \cite{YKI04,YIK09} in stochastic gradient algorithms to the current case of Langevin dynamics.

Most existing literature analyzes stochastic
approximation algorithms for tracking a parameter that evolves
according to a ``slowly time-varying'' sample path of a
continuous-valued process so that the parameter changes by
small amounts over small intervals of time. When the rate
of change of the underlying parameter is slower than the
adaptation rate of the stochastic approximation algorithm
(e.g., a slow random walk), the mean square tracking error can
be analyzed as in \cite{BMP90,KY03,SK95,Mou98}. In comparison,
our analysis covers the case where the global optimum
evolves with discrete jumps that can be arbitrarily large in
magnitude on short intervals of time. Also, the jumps can occur
on the same time scale as the speed of adaptation of the
stochastic approximation algorithm.
 Two-time scale and singularly  perturbed jump Markov systems are
 studied in~\cite{YZ06}.

Finally, we mention that  \cite{TTV16,RRT17} study convergence of the Langevin dynamics stochastic gradient algorithm in a non-asymptotic setting. Although the setting in our paper is asymptotic, it is interest in future work to study the non-asymptotic setting. 

\section{Adaptive Non-reversible Diffusion Algorithms}
\label{sec:algorithms}

The key  idea behind the adaptive algorithms below is to
parametrize $\th$ by $\skew$; denote this as $\th(\skew)$. Then one can pose a stochastic optimization problem  to find the skew symmetric matrix $\skew^*$ to minimize $\Cost(\th(\skew))$. In this section we propose three adaptive algorithms to adapt $\skew$; a Hessian based algorithm (Algorithm 1), a SPSA algorithm (Algorithm 2), and a two-time scale SPSA algorithm (Algorithm 3). From a practical point of view, the SPSA algorithm is numerically efficient and yields results comparable to the more expensive Hessian based algorithm.
 
\subsection{Algorithm 1. Hessian Based Adaptive Diffusion}  Let $\step, \alpha$ be small non-negative fixed step sizes with $\alpha = o(\step)$. Let $e_i$ denote the unit vector with 1 in the $i$-th position. Then the algorithm is as follows:
\begin{subequations} 
  \begin{empheq}[box=\widefbox]{align}
    \th_{k+1} &= \th_k  - \step \, (I + \skew_k ) \,  \D_\th \cost_k(\th_k) + \sqrt{\step}  \sqrt{\frac{2}{\temperature}} \snoise_k \label{eq:alg1a} \\
    {\skew_{k+1}(i,j)} &= \skew_k(i,j) - \alpha \,\D^\p_\th \cost_k(\th_k)   \, \thd_k(i,j) \Big\vert_{\skew^-}^{\skew^+}, \quad i > j  \label{eq:alg1b}\\
   \skew_{k+1}(i,j) &= -\skew_{k+1}(j,i), \quad i < j , \qquad \skew_{k+1}(i,i) = 0  \label{eq:alg1c}\\
   \thd_{k+1}(i,j) &= \thd_k(i,j) - \step \,(I+ \skew_k) \, \D_\th^2 \cost_k(\th_k)\, \thd_k(i,j)
   - \step \, (e_j - e_i)^\p \, \D_\th \cost_k(\th_k)   \label{eq:alg1d}
 \end{empheq}
\label{eq:alg1}
 \end{subequations}
 Eq.(\ref{eq:alg1a}) is simply the non-reversible diffusion (\ref{eq:nonrev}) with injected noise $\snoise_k$. Recall $\temperature> 0$ is the inverse temperature parameter   and $S_k$ is a $\thdim \times \thdim$ skew symmetric matrix. 
 In (\ref{eq:alg1b}), $\skew_0$ is initialized to an arbitrary skew symmetric matrix.
The notation $ \bigr\vert_{\skew^-}^{\skew^+}$ indicates that the  estimate $\skew_k(i,j)$ is projected onto the closed interval  $[\skew^-,\skew^+]$ if the estimate
lies outside this region.

The  recursion (\ref{eq:alg1b}) 
can be viewed as a stochastic gradient algorithm with step size $\alpha$ to minimize the cost
$\Cost(\th(\skew))$ wrt $\skew$. Formally, the gradient
$$\nabla_{\skew(i,j)} \Cost =
[\nabla _{\th} \Cost]^\p  \, \frac{d\th}{d \skew(i,j)} $$
so that an estimate of the gradient
$\D_{\skew(i,j)} \cost(\th_k) $  is  $\D^\p_\th \cost_k(\th_k)   \, \thd_k(i,j)$ where
$\thd_k(i,j) = \frac{d\th}{d \skew(i,j)}$.

The third equation (\ref{eq:alg1c}) enforces that $\skew_k$ is skew symmetric, namely $\skew^\p_k = -\skew_k$.

The final recursion (\ref{eq:alg1d}) is obtained by taking the ``derivative'' of the first recursion with respect to $\skew$
 by defining the vector
$\thd_k(i,j) = \frac{d\th_k}{ d\skew(i,j)} \in \reals^\thdim$ by holding $\skew$ fixed. Then
differentiating (\ref{eq:alg1a})  wrt $\skew(i,j)$ yields
$$ \thd_{k+1}(i,j) = \thd_k(i,j) - \step \,(I+ \skew_k) \, \frac{\partial}{\partial \skew(i,j)} \D_\th \cost_k(\th_k) - \step \frac{d \skew_k}{d \skew(i,j)}\, \D_\th \cost_k(\th_k)
$$
and
$$ \frac{\partial}{\partial \skew(i,j)} \D_\th \cost_k(\th_k) = \D_\th^2 \cost_k(\th_k) \, \frac{d\th_k}{d \skew(i,j)} , \qquad
\frac{d \skew_k}{d \skew(i,j)} =( e_j -  e_i)$$
Note that (\ref{eq:alg1d})  involves the Hessian $\D_\th^2 \cost_k(\th_k)$.

Formally, 
the process $\thd_k$ in (\ref{eq:alg1d})  is interpreted as the derivative $(d  / d\skew) \th |_{\th=\th_k}$.
This derivative process  is defined in the mean square
sense as in \cite[p. 1406]{KY95}:
$$\lim_{\Delta\to 0} \E \left| \thd - \frac{\th({\skew+\Delta}) -\th(\skew)}
{\Delta} \right|^2 =0.$$

In summary, the  adaptive non-reversible diffusion algorithm (\ref{eq:alg1})
is given by cross
coupling  two stochastic gradient algorithms (first and second recursion)
along with the derivative update of $\th$ with respect to the parameter
$\skew$
(final recursion).   Note that when $\skew$  is a fixed constant, $\skew^+=\skew^-=\skew$, then algorithm \eqref{eq:alg1} reduces to  the non reversible diffusion algorithm
(\ref{eq:nonrev}).

\subsection*{Main Convergence  Result (Informal)}
Since Algorithm 1  uses a constant step size (as opposed to a decreasing step size),  the appropriate notion of convergence is weak convergence \cite{KY03,EK86,Bil99}. Recall that
weak convergence is a function space generalization of convergence in distribution of random variables.
The assumptions and  main result will be stated formally and proved in Sec.\ref{sec:weak}. Here we give a heuristic statement.
We will show that the sequence of estimates $\{\th_k\}$ generated by the Algorithm 1 converges weakly to a non-reversible accelerated diffusion with the optimal skew symmetric matrix.

As is typically done in weak convergence analysis,
we first represent the sequence of estimates $\{\th_k\}$ generated by Algorithm~1 as a continuous-time process. This is done by constructing the continuous-time trajectory  via piecewise
constant interpolation.
Let $\horizon$  denote a positive real number which denotes the finite time horizon.
For
 $t   \in [0,\horizon]$,
define the continuous-time piecewise constant interpolated processes parametrized by the step size
$\step$ as
 $$ \th^\e(t)= \th_k \; \text{ for } \ t\in [\e k, \e k+ \e)$$

 We can now state our main result
 
 \begin{result} Under suitable assumptions (Sec.\ref{sec:weak}),  the interpolated processes $(\th^\e\cd,\thd^\e\cd,  \skew^\e\cd)$ converges weakly to
   $(\th\cd, \thd\cd, \skew\cd)$ such that the limit satisfies the following system of equations
   \begin{equation}
     \label{eq:1}
     \begin{split}
       d\th &=  (I + \skew) \,\nabla \Cost(\th) + \sqrt{\frac{2}{\temperature}} \, d\bm  \\
       \frac{d}{dt} \thd(i,j) &= (I + \skew)\, \nabla^2 \Cost(\th) \,\thd(i,j) - (e_j - e_i)^\p\, \nabla C(\th) \\
        \frac{d}{dt} \skew(i,j) &= \nabla \Cost(\th) \, \thd(i,j) \Big\vert_{\skew^-}^{\skew^+}, \quad i > j 
     \end{split}
   \end{equation}
 where $\bm\cd$ is $\thdim$-dimensional Brownian motion. \qed  
\end{result}

The most important takeaway from the above result is that the skew symmetric matrix $\skew$ satisfies the projected ordinary differential equation (ODE).
\beq \dot\skew(i,j) = -\frac{d}{d\skew(i,j)} \Cost(\th(\skew)),
\quad  \skew \in (\skew^-,\skew^+)  \label{eq:hyperode}\eeq
%The meaning \cite{KY03} is that $\dot \theta = - \Cost'_\th(\th)  {d\th \over d\skew}
 %    $ for $\skew \in (\skew^-,\skew^+)$, with $\skew^-$ and $\skew^+$ being absorbing points in %that if
%     $\Cost'_\th(\th)  {d\th \over d\skew} \le \skew^-$
 %    and
 %    $\Cost'_\th(\th)  {d\th \over d\skew}\ge \skew^+$, respectively.
Note that  (\ref{eq:hyperode}) implies that the adaptive algorithm for adjusting $\skew$ is a gradient decent method. By the weak convergence, $\skew_k$ will spend nearly all of the time in an arbitrarily small neighborhood of the local minima of $\E\{ \cost_k(\th(\skew))\}$, which is consistent with our motivation for the adaptive non-reversible diffusion  Algorithm 1.
 
\subsection{Algorithm 2.  SPSA based Adaptive Diffusion}
An issue with Algorithm~1 is that the computational cost  is
$O(\thdim^4)$ which is excessive for large $\thdim$; this computational cost is due to the update (\ref{eq:alg1d}) which is $O(\thdim^2)$ for each $i,j$. Also evaluating the 
Hessian
$\D_\th^2 \cost_k(\th)$ 
can be difficult in  some stochastic optimization problems.    Examining (\ref{eq:alg1}), we see that the Hessian arises as a by-product of evaluating the gradient estimate   $\Ds \cost_k(\th_k)$. Below we propose a finite difference evaluation of $\Ds \cost_k(\th_k)$; this does not involve the Hessian. But a naive
 evaluation of the finite difference approximation to gradient
 $\Ds \cost_k(\th_k)$
would require $2\thdim^2$  evaluations (simulations) of the cost; namely evaluate $\cost_k(\th(S + e_{ij}))$ and $\cost_k(\th(S-e_{ij}))$ for each  component $(i,j)$ of $\skew$.  The main idea below is to evaluate this gradient estimate using the  SPSA (simultaneous perturbation stochastic approximation)
algorithm \cite{Spa03}. The SPSA algorithm picks two random matrices $S_k + \stepa \Delta_k$ and $S_k - \stepa \Delta_k$ to evaluate
 $\Ds \cost(\th_k)$  and  therefore requires only 2 evaluations (simulations) of the cost $\cost_k(\cdot)$.

Let $\step, \stepa, \alpha$ be small non-negative fixed step sizes.
 We propose the following SPSA based algorithm that does not require computation of the Hessian
 \begin{equation} \begin{split}
        \th^+_{k+1} &= \th_k ^+ - \step \, (I + \skew_k + \stepa \,\Delta_k) \,  \D_\th \cost_k(\th_k^+) 
        + \sqrt{\step}  \sqrt{\frac{2}{\temperature} }\snoise_k \\
  {       \th^- _{k+1}} &= \th_k^-  - \step \, (I + \skew_k - \stepa \,\Delta_k) \,  \D_\th \cost_k(\th_k^-)  
        + \sqrt{\step}  \sqrt{\frac{2}{\temperature} }\snoise_k \\
     \skew_{k+1}(i,j) &= \skew_k(i,j) - \alpha   \frac{ \cost_k(\th^+) - \cost_k(\th^-)}{2\, \stepa\, \Delta_k(i,j)} 
   \end{split}
 \end{equation}
 Here the elements of the matrix $\Delta_k$ are simulated as follows: $\Delta_k(i,i) = 0$
\beq  \Delta_k(i,j) = \begin{cases} -1 & \text{ w. p. } 0.5 \\ 1 & \text{ w. p } 0.5 
\end{cases} , \quad i > j, \qquad \Delta_k(i,j) = - \Delta_k(j,i),  \quad  i< j  \label{eq:rand_direction}
\eeq
Note  $\skew_k + \stepa \,\Delta_k$ and  $\skew_k - \stepa\, \Delta_k$ are skew symmetric matrices by construction.

Algorithm 2 has computational cost of $O(\thdim^2)$ at each iteration.

\subsection{Algorithm 3. Two time scale SPSA Adaptive Diffusion}
Algorithm 2 discussed above simultaneously evaluates the gradient and updates the estimates in one time step. In comparison, we now construct a 
 two-time scale algorithm that proceeds as follows:

Run the following recursion  on the slow time scale $k=1,2,\ldots$, 
\beq
\th_{k+1} = \th_k  - \step \, (I + \skew_k ) \,  \D_\th \cost_k(\th_k)  + \sqrt{\step}  \sqrt{\frac{2}{\temperature} }\snoise_k \label{eq:step1_adaptive}
\eeq
and simulate $\Delta_k$ according to (\ref{eq:rand_direction}).
Then for each $k$, run multiple steps $n=0,\ldots,N$ on the fast time scale  to evaluate the estimate $D_k$ of the  gradient of $\cost_k(\th)$  wrt $\skew_k$: Initialize
$\th^+_0 = \th^-_0 =\th_k$ and
\begin{equation} \begin{split}
        \th^+_{n+1} &= \th_{n}^+ - \step \, (I + \skew_k + \stepa \,\Delta_k) \,  \D_\th \cost_n(\th_{n}^+) 
        + \sqrt{\step}  \sqrt{\frac{2}{\temperature} }\snoise_n \\
        \th^- _{n+1}&= \th_n^-  - \step \, (I + \skew_k - \stepa \,\Delta_k) \,  \D_\th \cost_n(\th_n^-)  
        + \sqrt{\step}  \sqrt{\frac{2}{\temperature} }\snoise_n \\
     D_{n+1}(i,j) &= D_n(i,j) +    \frac{ \cost_n(\th^+) - \cost_n(\th^-)}{2\, \stepa\, \Delta_n(i,j)} 
   \end{split}
   \label{eq:step2_adaptive}
 \end{equation}
Finally,  update $\skew$ in (\ref{eq:step1_adaptive}) on the slow time scale as 
 \beq
 \skew_{k+1}(i,j)  = \skew_k(i,j) - \alpha \,D_N(i,j) \eeq

Note that in the special case where (\ref{eq:step1_adaptive}) is omitted and (\ref{eq:step2_adaptive}) is run for one step, Algorithm 3 specializes to Algorithm 2.

\section{Numerical Examples. Adaptive  KL Divergence and Bayesian Learning}
This section   compares the performance of our proposed  non-reversible diffusion algorithms (Algorithms 1 2 and 3) to the classical Langevin algorithm in  numerical examples. We present  with a low dimensional KL divergence/Bayesian learning problem ($\thdim=2$) and  then a larger $\thdim=10$ dimensional problem.
In both cases, we show that Algorithms 1 and 2 converge faster  than the accelerated non-reversible diffusion (\ref{eq:nonrev}); which in turn converges faster than the classical Langevin~\eqref{eq:sgl}.

\subsection{Estimating KL Divergence}

The aim  is to  use the adaptive algorithms proposed above to explore and reconstruct high value regions of the KL divergence of the posterior.  As will be discussed below, a  special case of this setup is Bayesian learning discussed in \cite{WT11},
  where the algorithms explore high probability regions of the posterior distribution.

Let $\thtrue \in \reals^\thdim$ denote a true parameter value (which is unknown to the algorithm).  Let $\th \in \reals^\thdim$ denote a random variable with known prior distribution $\pdf(\th)$.
 A sequence of independent observation random variables\footnote{In this section, we use upper case $\Obs$  for random variables and lower case $\obs$  for their realization.}
 $\{\Obs_k\} $,  are generated from a known  likelihood $\pdf(\obs|\thtrue)$.
 The KL divergence of a sequence of $\horizon$ observations is
 \begin{equation}
     \label{eq:KL}
   \KL(\th^o,\th) = -\E_{\th^0}\{\log \frac{\pdf(\th|\Obst)}{\pdf(\th^o|\Obst)} \}
 \end{equation}
Given the observation sequence $\{\obs_k\}$, suppose we use the proposed  algorithms on expected cost
 \begin{equation}
   \label{eq:KLcost}
   \Cost(\th) = - \E_{\th^o}\{ \log \pdf(\th, \Obst) \} = \int \log\pdf (\th,\obst)\, \pdf(\obst|\th^o)\, d\obs_1,\ldots d\obs_\horizon
 \end{equation}
 A naive implementation of the unbiased gradient estimate is $\nabla_\th \pdf(\th,\obst)$; this uses
 batches of observations of length $\horizon$ from the sequence $\{\obs_k\}$.
 However, 
since the observations $\obs_k$ are iid, we can instead use a single observation $\obs_k$ at each time $k$ as an  unbiased  sample path gradient of the cost:
 \begin{equation}
   \label{eq:grad1}
  \D_\th \cost_k(\th_k) =   - \nabla_\th \log  \pdf(\th_\dtime) - \horizon\, \nabla_\th \log \pdf(\obs_\dtime|\th_\dtime)
\end{equation}
With this setup, suppose the Langevin   dynamics or any of the proposed algorithms  above, are run  on the observation sequence
$\{\obs_k\}$,  generated from the likelihood $\pdf(\obs|\thtrue)$.
Then, clearly  the algorithms asymptotically generate samples $\{\th_k\}$   from the stationary distribution (\ref{eq:stationary1}), namely
\[  \belief(\th) \propto \exp(-\Cost(\th)) \propto  \exp( \KL(\th^o,\th))
\]
where the proportionality constant involves terms independent of $\th$.

 To summarize, the Langevin dynamics algorithm and non-reversible diffusion algorithms
 (Algorithms 1, 2 and 3) operating on observations $\{\obs_k\}$ can be used  with gradient estimate $\D_\th \cost_k(\th_k)$  in (\ref{eq:grad1}) to estimate the KL divergence.
 Specifically if the empirical histogram $\hist(\th)$ is constructed from the samples
 $\{\th_k\}$ generated by the various algorithms, then $\log \hist(\th) \propto
 \KL(\th^o,\th)$.

\subsubsection*{Remark. Bayesian Learning}
Bayesian learning described in~\cite{WT11} is a special case of the above setup. It  deals with
exploring
high probability regions of  a  posterior density.

The setup in~\cite{WT11} is as follows:
Suppose $\obst$ is a \textit{fixed realization} generated from $\pdf(\obs|\th)$.  Then \[\bar{\Cost}(\th) \defn - \log \pdf(\th,\obst) \] is a deterministic cost. This is unlike the cost $\Cost(\th)$ in~\eqref{eq:KLcost} which involves 
the sequence of random variables $\Obst$ and an expectation. Clearly 
the sample path cost $\D_\th \cost_k(\th_k) $ evaluated in (\ref{eq:grad1}) for $k\in 1,\ldots,\horizon$  is a noisy unbiased estimate of $\bar{\Cost}(\th)$.

 Suppose the Langevin   dynamics algorithm or any of the adaptive algorithms proposed above, are run  on the augmented dataset\footnote{In \cite{WT11} this is termed as running the algorithms on multiple sweeps of $\obst$. Also \cite{WT11} uses a decreasing step size algorithm.}  $\augmented = \obst, \obst, \ldots,$. Note the augmented dataset comprises multiple repetitions of $\obst$.
Then  the algorithms asymptotically generate samples $\{\th_k\}$   from the stationary distribution (\ref{eq:stationary1}), namely
\[  \belief(\th) \propto \exp(- \bar{\Cost}(\th)) = \pdf(\th|\obst)
  \]
  To summarize, the Langevin dynamics algorithm and non-reversible diffusion algorithms
  (Algorithms 1, 2 and 3) operating on augmented dataset $\augmented$ can be used  with gradient estimate $\D_\th \cost_k(\th_k)$  in (\ref{eq:grad1}) to perform Bayesian learning.
  That is, the algorithms  construct a non-parametric estimate of the posterior distribution $\pdf(\th|\Obs)$  from  the empirical density $\hist(\th)$ by using the  iterates $\{\th_k\}$ generated by the algorithms.

\subsection{Example  1. Bayesian Learning $\thdim = 2$}
Here we consider the case $\thdim = 2$, $\th = [\th(1), \th(2)]^\p$,
\begin{equation} \label{eq:2dim}  \begin{split}
            &\obs_k  \sim \frac{1}{2}\normal(\theta(1), \sqrt{2}) + \frac{1}{2}\normal(\theta(1) + \theta(2), \sqrt{2}) \\
        &    \theta(1) \sim \normal(0, \sqrt{10}), \quad  \theta(2) \sim \normal(0, 1)
          \end{split}
        \end{equation}
  
For  true parameter value  $\thtrue=[0, 1]^\p$, it can be verified that
the
objective  $-\Cost(\th)$ is non-concave in $\th$ and has two maxima at
$\th = [0, 1]^\p$ and $\th=[1, -1]^\p$.

\begin{figure}[h] \centering
  \begin{subfigure}{.45\textwidth}
    \includegraphics[scale=0.5]{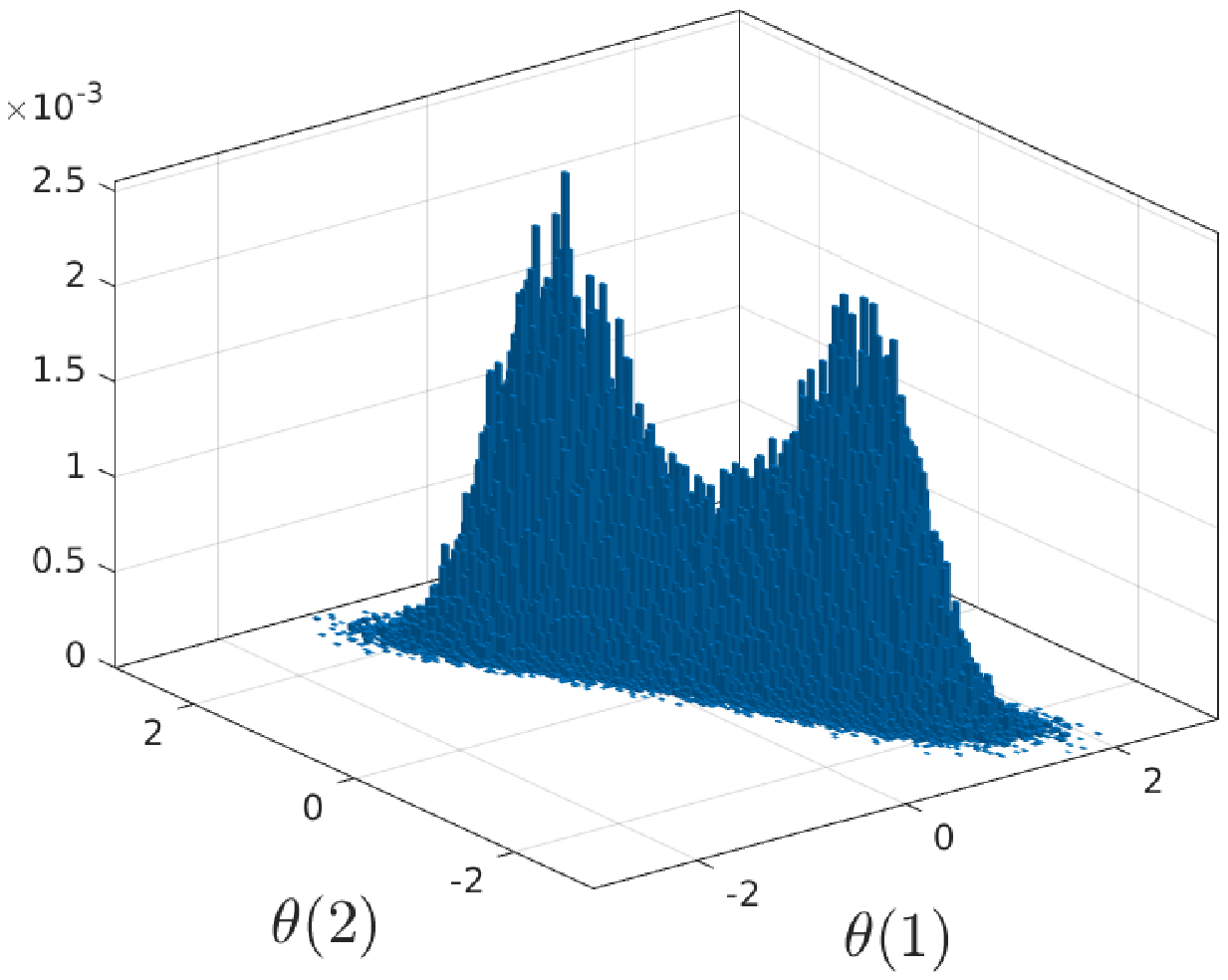}
   % \caption{Metropolis Hastings simulation of posterior distribution}
  \end{subfigure}
  \begin{subfigure}{.45\textwidth}
    \includegraphics[scale=0.5]{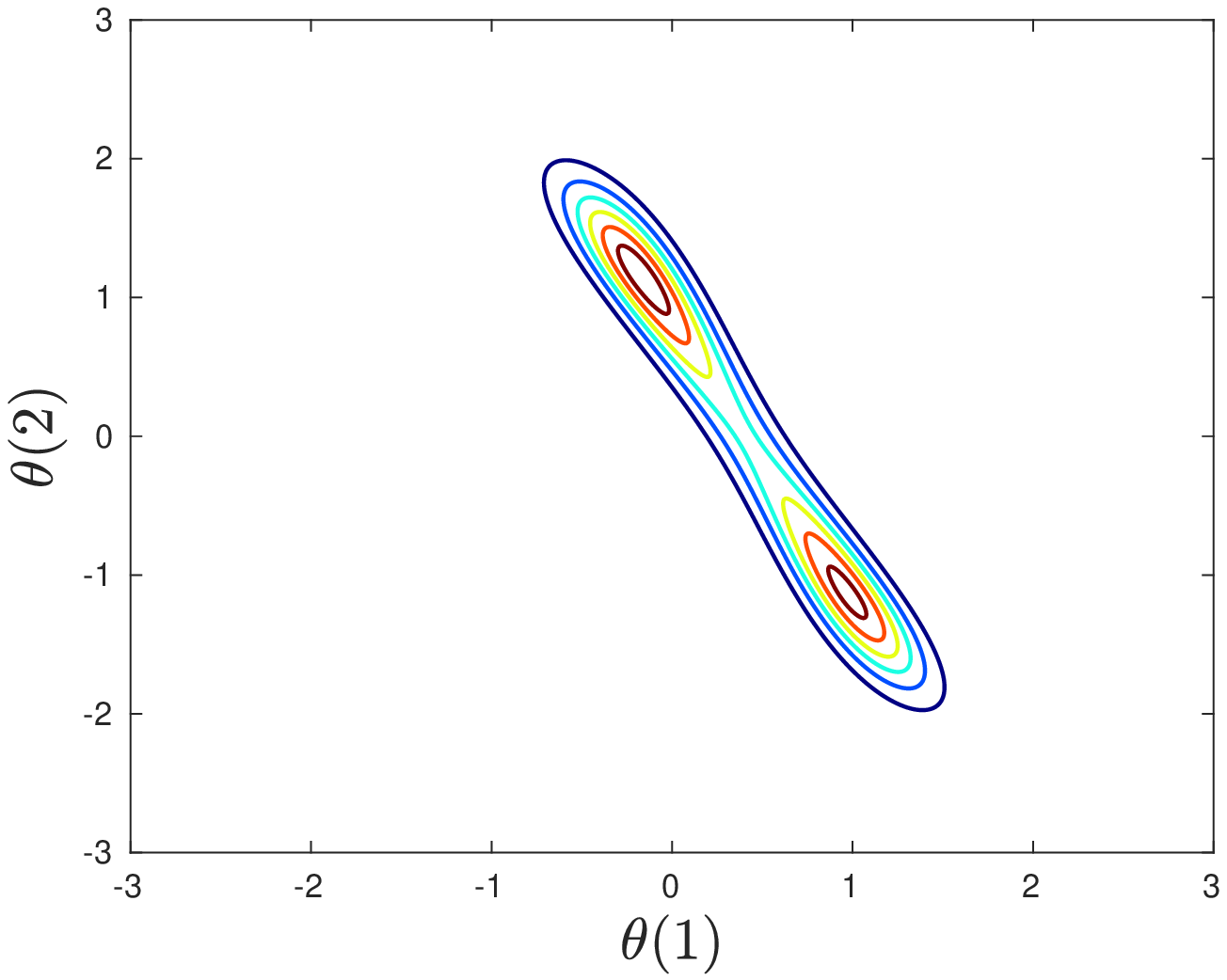}
   % \caption{Metropolis Hastings simulation of posterior distribution}
  \end{subfigure}
  \caption{Metropolis Hastings simulation of posterior distribution $\pdf(\th|\obs_1,\ldots,\obs_{\horizon})$, $\horizon=100$.}
  \label{fig:mh0}
\end{figure}

To illustrate the posterior $\pdf(\th|\obs_1,\ldots,\obs_\horizon)$ visually, Figure \ref{fig:mh0} plots the empirical density and contours  of $\pdf(\th|\obs_1,\ldots,\obs_\horizon)$,
$\th \in \reals^2$, for $\horizon = 100$ using the Metropolis Hastings algorithm.

The augmented dataset $\augmented$ was generated as $1000$ repetitions of $\obs_1,\ldots,\obs_{100}$; so $\augmented$ has $10^5$ points.
We ran the Langevin dynamics algorithm, accelerated algorithm and Algorithms 1, 2 and 3 with $\temperature = 1$ over augmented dataset $\augmented$
for 30 independent trials each with initial condition
$\theta_0 = [4,4]^\p$. Each trial has a different sample path of the  injected noise $\{\snoise_k\}$.  The $2\times 2 $ skew symmetric matrix was initialized as $\skew_0 = \begin{bmatrix} 0 & -s \\ s & 0 
\end{bmatrix}
$ where $s \sim \normal(0,1)$.

\begin{figure}[h] \centering
  \begin{subfigure}{.45\textwidth}
    \includegraphics[scale=0.5]{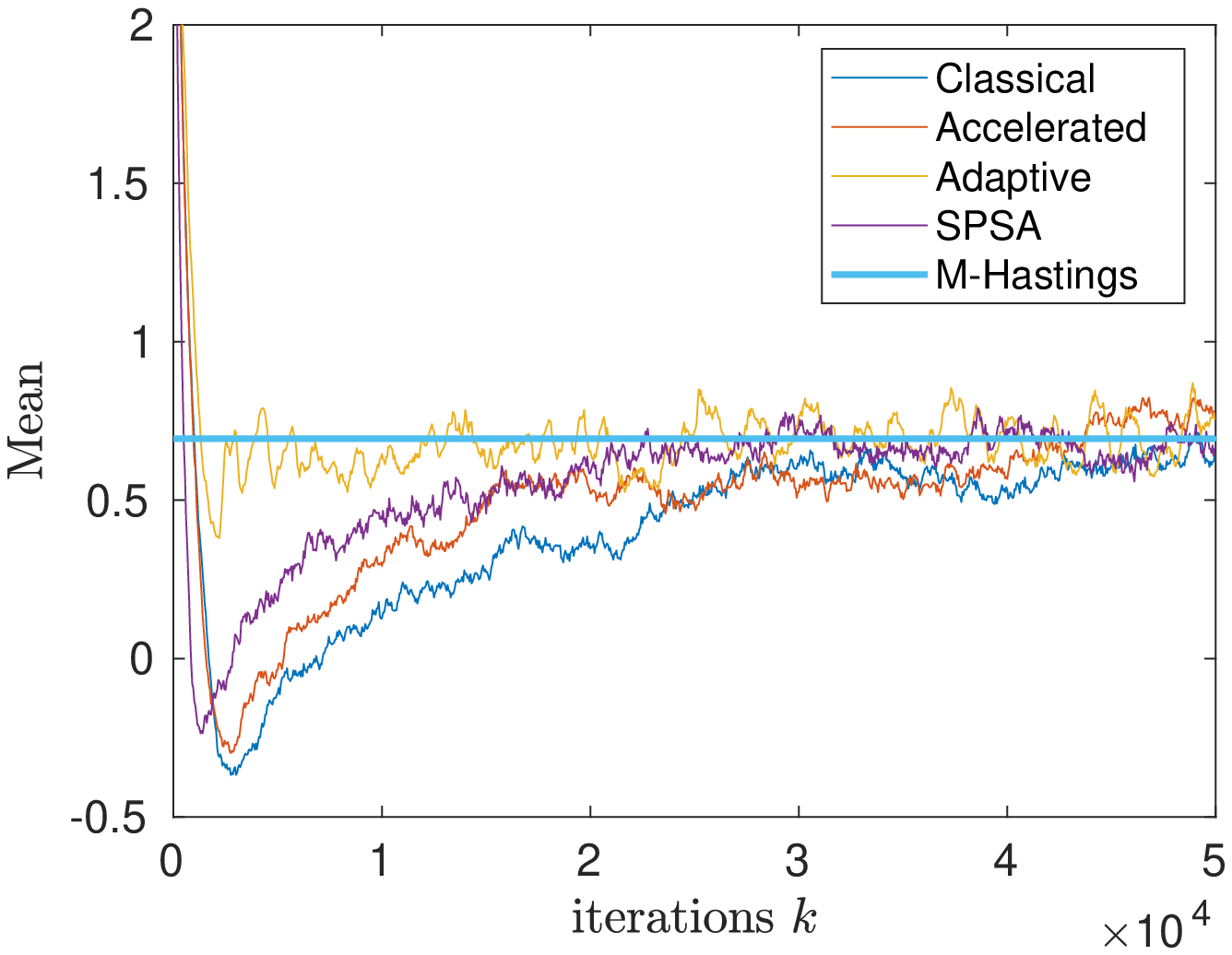}
 \caption{Posterior mean for $\th(1)$}
  \end{subfigure}
  \begin{subfigure}{.45\textwidth}
    \includegraphics[scale=0.5]{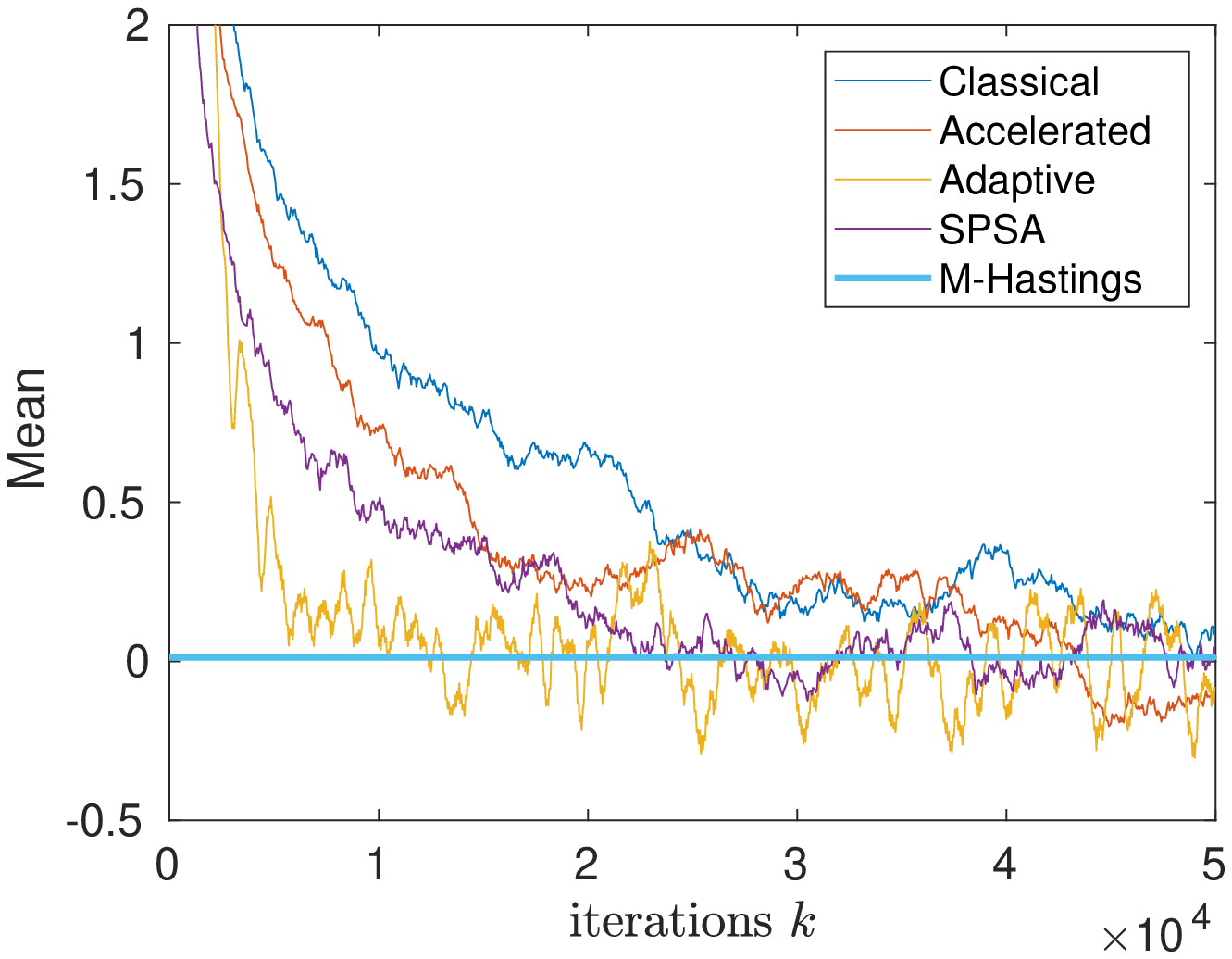}
 \caption{Posterior mean for $\th(2)$}
  \end{subfigure}
  \caption{$\thdim=2$. Comparison of posterior means $\E\{\th(i)|\obs_1,\ldots\obs_{100}\}, i=1,2$ versus
    iterations for various algorithms}
  \label{fig:postmean}
\end{figure}

Figure \ref{fig:postmean} displays the estimated posterior means $\E\{\th(i)|\obs_1,\ldots,\obs_{100}\}$, $i=1,2$.
As can be seen from Figure \ref{fig:postmean}, Algorithms 1, 2, and 3 converges faster than the accelerated algorithm,  which in turn converges faster than the classical Langevin.

\subsection{Example  2. Bayesian Learning $\thdim = 10$}
\begin{align*}
    \th(i) &\sim \normal(\mu_i, \sigma^2_i),  \quad
    \mu_i \sim \uniform[-2, 2], \;
    \sigma^2_i \sim \uniform[1, 10], \qquad i \in \{1, \dots \thdim\}, \\
   \obs_k &\sim \frac{1}{2}\normal(\sum_{i=1}^{\thdim/2}\th(i), \sqrt{2}) + \frac{1}{2}\normal(\sum_{j=\frac{\thdim}{2}+ 1}^\thdim\th(j), \sqrt{2})
\end{align*}
As in the previous example the aim is to reconstruct the posterior
$\pdf(\theta|\obs_1,\ldots,\obs_{100})$.

First, the Metropolis Hastings algorithm was used to generate samples from the posterior. We view the estimates from the Metropolis Hastings  as the ground truth.

Next we implemented the classical Langevin algorithm, accelerated algorithm and adaptive algorithms.
In the accelerated algorithm and Algorithms 1, 2,   the skew symmetric matrix  $\skew_0$ was initialized as a tri-diagonal   matrix with elements above the diagonal chosen as $\normal(0,1)$ random variables, and elements below the diagonal chosen as the negative of these.
The augmented dataset $\augmented$ was generated as $10^4$ repetitions of $\obs_1,\ldots,\obs_{100}$; so $\augmented$ has $10^6$ points.
Each algorithm was run for  50 independent trials with step sizes
$\step=10^{-4}$, $\alpha = 10^{-4}$.

\begin{figure}[h] \centering
  \begin{subfigure}{.45\textwidth}
    \includegraphics[scale=0.5]{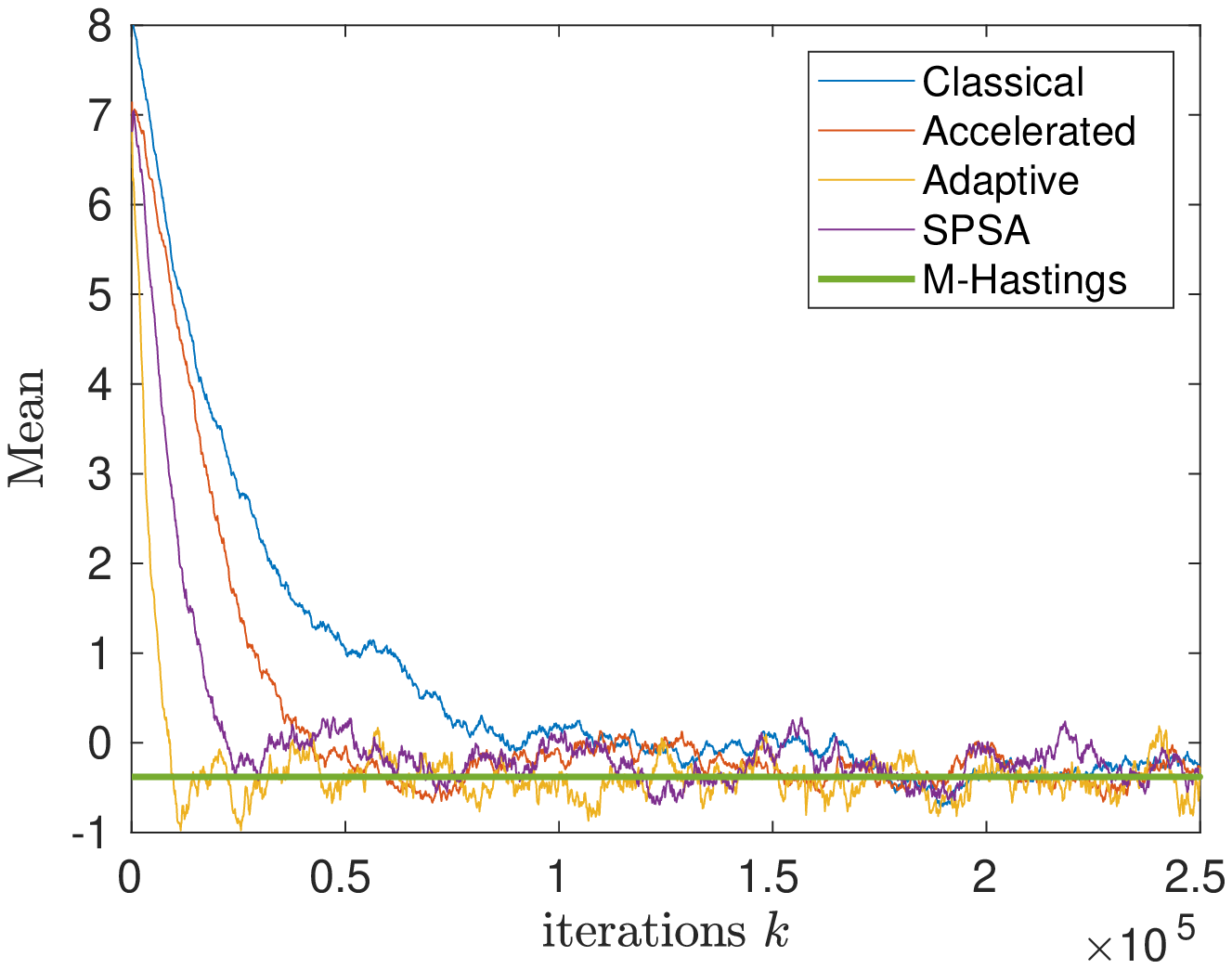}
 \caption{Posterior mean for $\th(1)$}
  \end{subfigure}
  \begin{subfigure}{.45\textwidth}
    \includegraphics[scale=0.5]{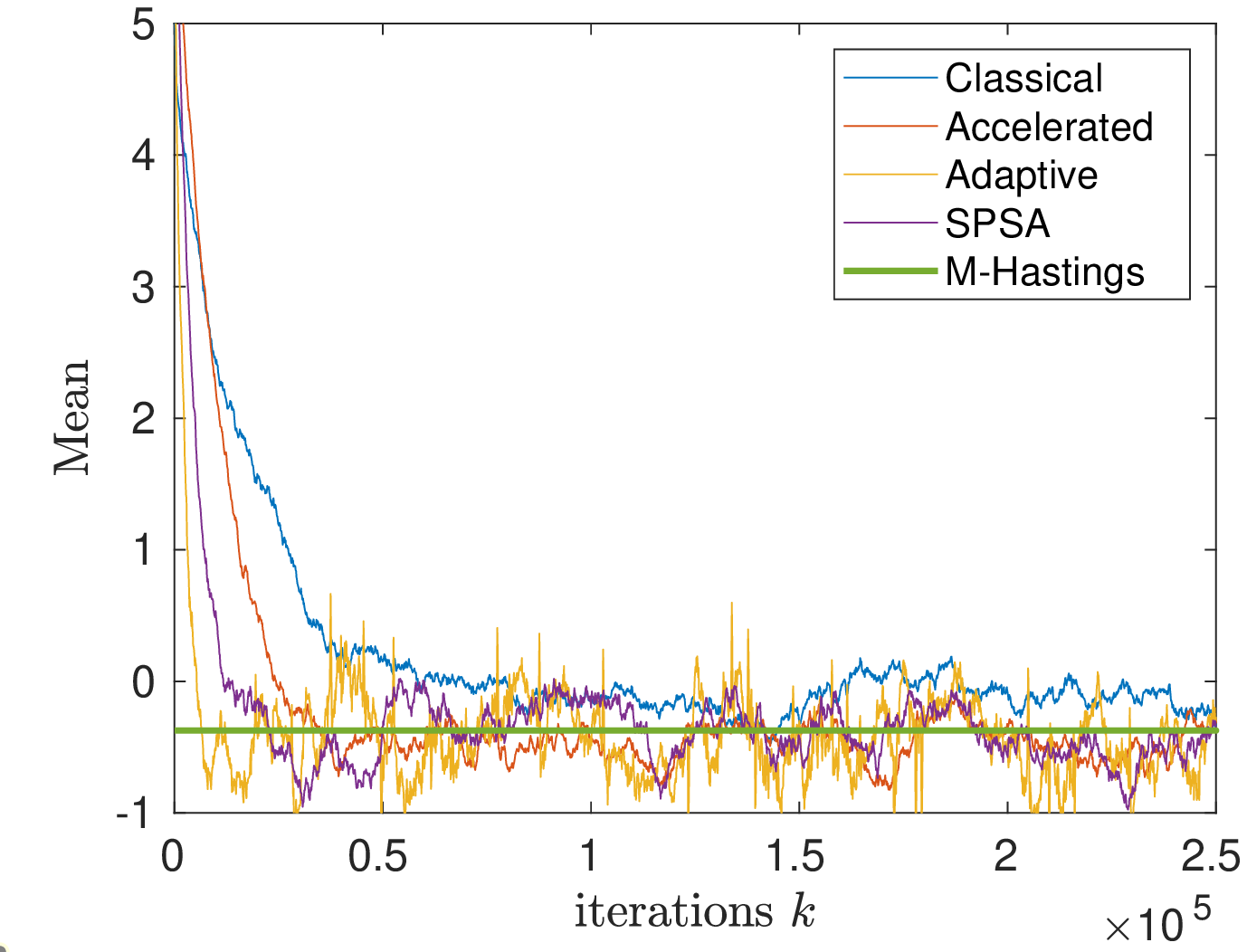}
 \caption{Posterior mean for $\th(2)$}
\end{subfigure}
 \begin{subfigure}{.45\textwidth}
    \includegraphics[scale=0.5]{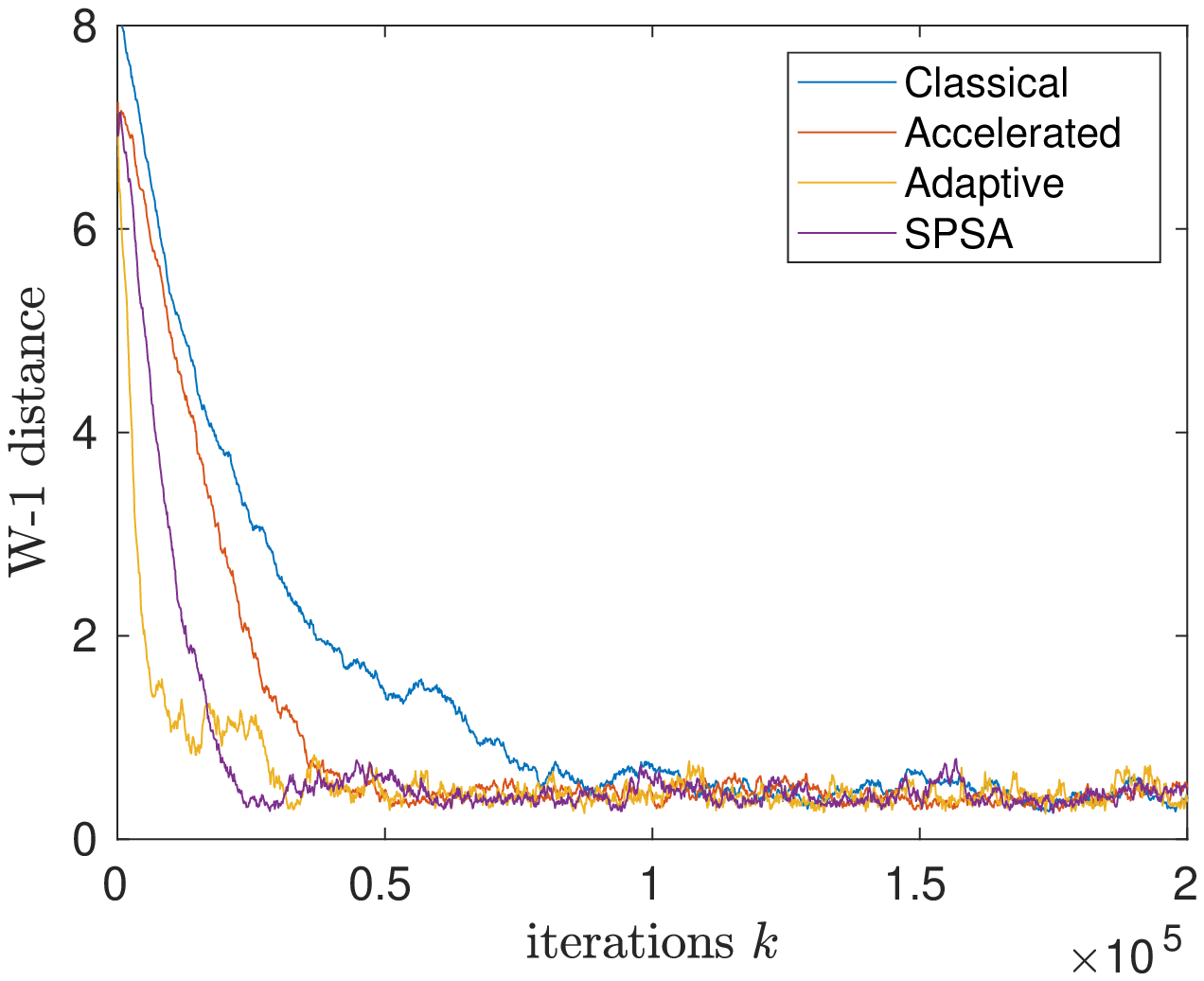}
 \caption{Wasserstein-1 distance for first marginal}
  \end{subfigure}
  \begin{subfigure}{.45\textwidth}
    \includegraphics[scale=0.5]{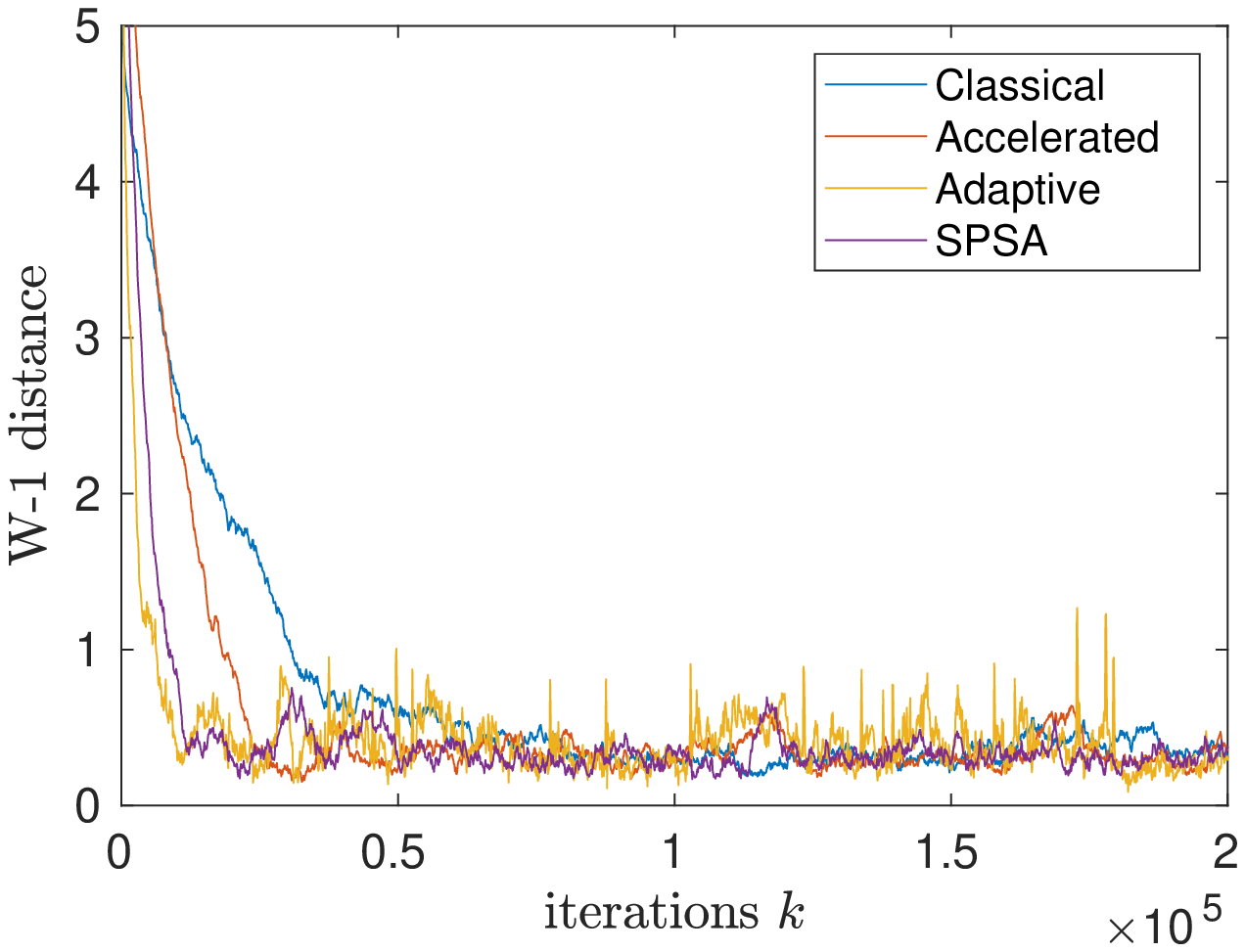}
 \caption{Wasserstein-1 distance for second marginal}
  \end{subfigure}
  \caption{$\thdim=10$. Comparison of posterior means $\E\{\th(i)|\obs_1,\ldots\obs_{100}\}$ and Wasserstein 1-distances
    $d(i)$ for the first two marginals $i=1,2$  versus
    iterations for various algorithms}
  \label{fig:postmean2}
\end{figure}

The posterior   $\pdf(\th|\obs_1,\ldots,\obs_{100})$ is a $10$-variate distribution
Figure \ref{fig:postmean2} shows the posterior mean estimates of the first two marginals, computed for the various algorithms. Also shown are the  $L_1$ distances of these marginals to that of the Metropolis Hastings algorithm.
The $L_1$ distance  (Wasserstein 1-metric)
for the first two marginals is
\beq
d(i)  = \int  | \hat{F}_i(\eth(i)) - F_i(\eth(i)) |\, d\eth(i), \quad i = 1,2
\label{eq:L1}
\eeq
where $F_i$ is the cumulative distribution of marginal $i$ constructed via Metropolis Hastings (ground truth) and
$\hat{F}_i$ is the empirical cumulative distribution constructed by the Langevin or adaptive algorithm.

The $L_1$ distance is more appropriate for our purposes than the Kolmogorov-Smirnov distance since
typically the constant or proportionality $\temperature$ is not known and
so the regions of support of the empirical cdfs can vary substantially.

 % \subsection{Example 3.  Tracking time evolving  KL divergence}

%   An important feature of the algorithms proposed in this paper is that have a fixed step size (as opposed to a decreasing step size). So the  Lagenvin dynamics and adaptive algorithms can track and reconstruct time-evolving objective functions.
%  Here we illustrate via numerical examples 
% how well the algorithms reconstruct the  KL divergence when  the underlying true parameter
% $\th^o$ (unknown to the algorithm) jump changes at an unknown time. In Sec.\ref{sec:track} we will formally analyze the tracking properties of the adaptive algorithms when the true parameter evolves randomly according  to a slow Markov chain.   

{\em Acknowledgement}. The above simulations were done Cornell graduate student  Omer Serbetci.

%\subsection{Example 3. Feed Forward Neural Network}

\section{Weak Convergence Analysis} \label{sec:weak}

\section{Non-stationary Global Optimization and Tracking Analysis} \label{sec:track}
Our next main result concerns estimating  a time evolving global
minimum in a non-stationary global stochastic optimization problem. Alternatively, we use to use non-reversible diffusion based algorithms to explore and track a time evolving expected 
Since we are  estimating (tracking) a time evolving  global minimum/cost, we first  give a model for the evolution.
Below, the Markov chain $\{\mc_k\}$ will be used as a hyper-parameter to model the evolution
of the global minimum.  By hyper-parameter we mean that the Markov chain model is not known or used  by the algorithms. The Markov chain assumption is used  only for our convergence analysis
to determine how well does our proposed algorithm estimates (tracks) a global minimum/expected cost that jump changes
(evolves) according to an unknown Markov chain.

\subsection{Non-stationary Stochastic Optimization  Problem}
In this section, we treat the problem minimization of an objective function in which the objective function is randomly changing within a finite set. Effectively, instead of one objective function, we have a finite number of objective functions to deal with.
For the reason of mathematical convenience, we assume that the  random changing behavior is modeled by  a ``slow''  Markov chain $\{\state_k\}$ on the finite state space  $\statespace=\{1,\dots,X\}$ and the one-step transition probability   $I +\mcstep Q$. Here $\mcstep>0$ is a small parameter and $Q=(q_{ij})$ is a generator of a continuous-time Markov chain so that $q_{ij}\ge 0$ for $i\not =j$
and $\sum_j q_{ij}=0$ for each $i \in \statespace$.
We assume that $Q$ is irreducible (see \cite[p.23]{YZ13}).
For notational convenience, we have chosen the states of the Markov chain to take integer values. This is no loss of generality.

With the above setup, we carry out an optimization problem of the form
\beq
\begin{split}
\th_\mc \in \Th  &= \{ \argmin_{\th \in \Th \subset \reals^\thdim, \mc\in \statespace}  \cost(\th,\mc) \}  \\  \text { where }  &  \cost(\th ,\mc) =  \E_\mc\{ \cost(\th , \mc,\signal) \} = \E \{ \cost(\th , \mc,\signal) |\mc_k=\mc \},
 %= \int_Z \costsample(\th ,\signal)  P_{\mc_k}(d\signal)
 \end{split}
 \label{eq:stochopt}\eeq
 where $z$ is the observation. The above optimization is taken as conditional expectation conditioned on $\mc_k =\mc$. Thus in lieu of
 one objective function, we have $X$ objective functions. Thus equivalently, we are treating a time-varying tracking problem of tracking the time-varying minimizer.

 \bibliographystyle{IEEEtran}

\bibliography{$HOME/texstuff/styles/bib/vkm}

\end{document}